\documentstyle[aclap,epsbox]{article}

\title{\vspace{-0.5in}Robust Parsing Based on Discourse Information:\\
{\large {\bf Completing partial parses of ill-formed sentences \\
 on the basis of discourse information}}}
\author{Tetsuya Nasukawa \\
IBM Research, Tokyo Research Laboratory \\
1623-14, Shimotsuruma, Yamato-shi, Kanagawa-ken 242, Japan \\
{\tt nasukawa@trl.vnet.ibm.com}
}

\begin{document}

\maketitle
\vspace{-0.5in}
\begin{abstract}

In a consistent text,
many words and phrases are repeatedly used in more than one sentence.
When an identical phrase (a set of consecutive words)
is repeated in different sentences, the constituent words of those
sentences tend to be associated in identical modification patterns with
identical parts of speech and identical modifiee-modifier relationships.
Thus, when a syntactic parser cannot parse a sentence as a unified
structure, parts of speech and modifiee-modifier relationships among
morphologically identical words in complete parses of other sentences
within the same text provide useful information for obtaining partial
parses of the sentence.

In this paper, we describe a method for completing partial parses
by maintaining consistency among morphologically identical words within
the same text as regards their part of speech and their
modifiee-modifier relationship.
The experimental results obtained by using this method with technical
documents offer good prospects for improving the accuracy of sentence
analysis in a broad-coverage natural language processing system such as
a machine translation system.

\end{abstract}

\section{Introduction}

In order to develop a practical natural language processing (NLP)
system, it is essential to deal with ill-formed sentences that cannot be
parsed correctly according to the grammar rules in the system.
In this paper, an ``ill-formed sentence'' means one that cannot be
parsed as a unified structure.
A syntactic parser with general grammar rules is often unable to analyze
not only sentences with grammatical errors and ellipses, but also
long sentences, owing to their complexity.
Thus, ill-formed sentences include not only ungrammatical sentences,
but also some grammatical sentences that cannot be parsed as unified
structures owing to the presence of unknown words or to a lack of
completeness in the syntactic parser.
In texts from a restricted domain, such as computer manuals,
most sentences are grammatically correct.
However, even a well-established syntactic parser usually fails to
generate a unified parsed structure for about 10 to 20 percent of all
the sentences in such texts, and the failure to generate a unified
parsed structure in syntactic analysis leads to a failure in the output
of a NLP system.
Thus, it is indispensable to establish a correct analysis for
such a sentence.

To handle such sentences, most previous approaches
apply various heuristic
rules \cite{Jensen:1983,Douglas:1992,Richardson:1988}, including
\begin{itemize}
\item Relaxing constraints in the condition part of a grammatical rule,
such as number and gender constraints
\item Joining partial parses by using meta rules.
\end{itemize}
Either way, the output reflects the general plausibility of an analysis
that can be obtained from information in the sentence; however,
the interpretation of a sentence depends on its discourse, and inconsistency
with recovered parses that contain different analyses of the same phrase
in other sentences in the discourse
often results in odd outputs of the natural language processing system.

Starting from the viewpoint that an interpretation of a sentence must be
consistent in its discourse, we worked on completing incomplete parses
by using information extracted from complete parses in the discourse.
The results were encouraging.
Since most words in a sentence are repeatedly used in other
sentences in the discourse, the complete parses of well-formed sentences
usually provided some useful information for completing incomplete
parses in the same discourse.
Thus, rather than trying to enhance a syntactic parser's grammar rules
in order to support ill-formed sentences, which seems to be an endless
task after the parser has obtained enough coverage to parse general
grammatical sentences,
we treat the syntactic parser as a black box and
complete incomplete parses, in the form of partially parsed chunks
that a bottom-up parser outputs for ill-formed sentences,
by using information extracted from the discourse.

In the next section, the effectiveness of using information
extracted from the discourse to complete syntactic analysis of
ill-formed sentences.
After that, we propose an algorithm for completing incomplete
parses by using discourse information,
and give the results of an experiment on completing
incomplete parses in technical documents.

\section{Discourse information for completing incomplete parses}

\begin{figure*}[htbp]
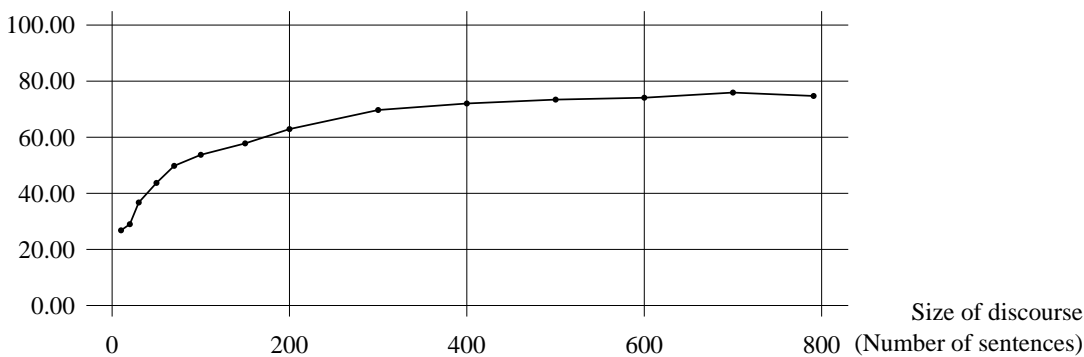

 \begin{center}
 \epsfile{file=Col-Freq.ps,width=150mm}
 \end{center}
\caption{Rate of finding identical or similar collocation patterns
in relation to the size of the discourse}
 \label{Figure-Size-and-MIW}
\end{figure*}

\begin{table*}[ht]
\caption{Frequency of morphologically identical words in computer
manuals}
\label{Table-Fre-of-MIW}
\begin{center}
{\small
\begin{tabular}{|c|c|c||c|c|} \hline
Part & \multicolumn{2}{c||}{Freq. of morph. identical words} &
\multicolumn{2}{c|}{Proportion of all content words} \\ \cline{2-5}
of & Two or more & Five or more & Total number of & Proportion \\
speech & times (\%) & times (\%) & appearances (words) & (\%)
 \\ \hline
Noun & 90.7 & 76.2 & 99047 & 59.8 \\ \hline
Verb & 94.9 & 83.6 & 35622 & 21.5 \\ \hline
Adjective & 88.9 & 71.0 & 16941 & 10.2 \\ \hline
Adverb & 85.9 & 68.8 & 4993 & 3.0 \\ \hline
Pronoun & 98.0 & 94.8 & 8911 & 5.4 \\ \hline \hline
Total & 91.6 & 78.0 & 165514 & --- \\ \hline
\end {tabular}
}
\end{center}
\end{table*}
In this section, we use the word ``discourse'' to denote
a set of sentences that forms
a text concerning related topics.
Gale \cite{Gale:1992b} and Nasukawa \cite{Nasukawa:1993} reported that
polysemous words within the same discourse have the same word sense with
a high probability (98\% according to \cite{Gale:1992b},) and the results
of our analysis indicate that most content words are frequently
repeated in the discourse, as is shown in Table \ref{Table-Fre-of-MIW};
moreover, collocation (modifier-modifiee relationship) patterns are also
repeated frequently in the same discourse, as is shown in
Figure \ref{Figure-Size-and-MIW}.
This figure reflects the analysis of structurally ambiguous phrases in a
computer manual consisting of 791 consecutive sentences for discourse
sizes ranging from 10 to 791 sentences.
For each structurally ambiguous phrase,
more than one candidate collocation pattern was formed by associating
the structurally ambiguous phrase with its candidate modifiees
\footnote{For example, in the sentence\\
{\it You can use the folder on the desktop,}\\
the ambiguous phrase, {\it on the desktop}, forms two candidate
collocation patterns:\\
 ``use --(on)-- desktop'' and ``folder --(on)-- desktop.''},
and a collocation pattern identical with or similar to each of these
candidate collocation patterns was searched for in the discourse.
An identical collocation pattern is one in which both modifiee and
modifier sides consist of words that are morphologically identical
with those in the sentence being analyzed, and that stand in an
identical relationship.
A similar collocation pattern is one in which either the modifiee or
modifier side has a word that is morphologically identical with the
corresponding word in the sentence being analyzed, while the other
has a synonym.
Again, the relationship of the two sides is identical with that in
the sentence being analyzed.
Except in the case where all 791 sentences were referred to as a
discourse, the results indicate the averages obtained by referring to
each of several sample areas as a discourse.
For example, to obtain data for the case in which the size of a
discourse was 20 sentences, we examined 32 areas each consisting of 20
sentences, such as the 1st sentence to the 20th, the 51st to the 70th,
and the 701st to the 720th.
Thus, Figure \ref{Figure-Size-and-MIW} indicates that a collocation
pattern either identical with or similar to at least one of the
candidate collocation patterns of a structurally ambiguous phrase was
found within the discourse in more than 70\% of cases, provided the
discourse contained more than 300 consecutive sentences.

On the assumption that this feature of words in a discourse provides a
clue to improving the accuracy of sentence analysis,
we conducted an experiment on sentences for which a
syntactic parser generated more than one parse tree,
owing to the presence of words that can be assigned to more than one
part of speech, or to the presence of complicated coordinate
structures, or for various other reasons.
If the constituent words tend to be associated in identical modification
patterns with an identical part of speech and identical
modifiee-modifier relationship when an identical phrase (a set of
consecutive words) is repeated in different sentences within the
discourse, the candidate parse that shares the most collocation patterns
with other sentences in the discourse should be selected as the correct
analysis.
Out of 736 consecutive sentences in a computer manual, the ESG parser
\cite{McCord:1991} generated multiple parses for 150 sentences.
In this experiment, we divided the original 736 sentences into two texts,
one a discourse of 400 sentences and the other a discourse of
336 sentences.
Of the 150 sentences with multiple parses, 24 were incorrectly analyzed
in all candidate parses or had identical candidate
parses; we therefore focused on the other 126
sentences.
In each candidate parse of these sentences, we assigned a score for each
collocation that was repeated in other sentences in the discourse
(in the form of either an identical collocation or a similar collocation),
and added up the collocation scores to assign a preference value to
the candidate parse.
Out of the 126 sentences, different preference values were assigned to
candidate parses in 54 sentences, and the highest value was
assigned to a correct parse in 48 (88.9\%) of the 54
sentences.
Thus, there is a strong tendency for identical collocations to be
actually repeated in the discourse, and when an identical phrase (a set
of consecutive words) is repeated
in different sentences, their constituent words tend to be associated in
identical modification patterns.

\begin{figure*}[ht]
\begin{quote}
\begin{quote}
{\footnotesize
\baselineskip=0.9\normalbaselineskip
\begin{verbatim}
((XXXX (COMMENT(CONJ   "as")
               (NP     (PRON*  "you" ("you" (SG PL))))
               (AUXP   (VERB*  "can" ("can" PS)))
               (VERB*  "see" ("see" PS)))
       (PUNC   ",")
       (VP     (NP     (PRON*  "you" ("you" (SG PL))))
               (AUXP   (VERB*  "can" ("can" PS)))
               (VERB*  "choose" ("choose" PS))
               (PP     (PP     (PREP*  "from"))
                       (QUANP  (ADJ*   "many" ("many" BS)))
                       (NOUN*  "topics" ("topic" PL))))
       (VP*    (INFCL  (INFTO  (PREP*  "to"))
                       (VERB*  "find" ("find" PS))
                       (COMPCL (COMPL  "")
                               (VERB*  "out" ("out" PS))
                               (NP     (PRON*  "what" ("what" (SG PL))))))
               (NP     (NOUN*  "information" ("information" SG)))
               (VERB*  "is" ("be" PS))
               (AJP    (ADJ*   "available" ("available" BS))
                ?      (PP     (PP     (PREP*  "about"))
                               (DETP   (ADJ*   "the" ("the" BS)))
                               (NP     (NOUN*  "AS/400" ("AS/400" (SG PL))))
                               (NOUN*  "system" ("system" SG)))))
       (PUNC   "."))  0)
\end{verbatim}
\baselineskip=1.8\normalbaselineskip}
\end{quote}
\end{quote}
\caption{Example of an incomplete parse obtained by the PEG parser}
\label{Fig-PEG-output}
\end{figure*}

Figure \ref{Fig-PEG-output} shows the output of the PEG parser
\cite{Jensen:1992} for the following sentence:
\begin{description}
\item[(2.1)]{\sl As you can see, you can choose from many topics to find
out what information is available about the AS/400 system.}
\end{description}
This is the 53rd sentence in Chapter 6 of a computer
manual \cite{IBM:1992},
and every word of it is repeatedly used in other
sentences in the same chapter, as shown in Table \ref{Table-DiscInfo-assign}.
\begin{table*}[ht]
\caption{Selecting POS candidates on the basis of discourse information}
\label{Table-DiscInfo-assign}
\begin{flushleft}
{\small
\begin{tabular}{|c| c c c c c c c c c c c c c|} \hline
 & As & you & can & see, & you & can & choose & from &
many & topics & to & find & out \\ \hline
Candidates & CJ & PN & N & N & PN & N & V & PP &
AJ & N & PP & N & PP \\
for the POS & AV &  & V & V &  & V &  &  & N &  &  & V & N \\
of each word  & PP &  &  &  &  & &  &  & PN &  &  & AV & PP \\
 &  &  &  &  &  & &  &  &  &  &  &  & V \\ \hline
 & As & you & can & see, &
\multicolumn{9}{l|}{{\it appears in sentences 39, 175.}} \\ \cline{6-14}
Phrases & CJ & PN & V & V &
\multicolumn{1}{|c}{you} & can & choose &
\multicolumn{6}{l|}{{\it appears in sentences 179.}} \\
  \cline{10-14}
repeated &  &  &  &  & \multicolumn{1}{|c}{PN} & V &
 V & & \multicolumn{1}{|c}{many} &
\multicolumn{4}{l|}{{\it appears in sentences 49.}} \\
\cline{6-9} \cline{11-14}
within the &  &  &  &  &  &  &  &  & \multicolumn{1}{|c|}{AJ} &
 \multicolumn{1}{c|}{topics} & & \multicolumn{2}{c|}{find out what} \\
 \cline{2-10} \cline{12-13}
discourse & \multicolumn{9}{l}{{\it appears in sentences 39,
140, 145, 160, 161 167 169...}} &
 N & \multicolumn{1}{|c}{to} & \multicolumn{1}{c|}{find} & \\
 \cline{2-11}
 & \multicolumn{10}{l}{{\it appears in sentences 236.}} &
 PP & \multicolumn{1}{c|}{V} & \\ \cline{2-13}
 & \multicolumn{11}{l}{{\it appears in sentences 32.}} &
\multicolumn{2}{c|}{V \ PP \ (PN)} \\ \hline
POS & CJ & PN & V & V & PN & V
& V & PP & AJ & N & PP & V & PP \\ \hline
\end{tabular}

\smallskip

\begin{tabular}{|c| c c c c c c c c |} \hline
 & what & information & is & available & about & the & AS/400 & system.
 \\ \hline
Candidates & AJ & N & V & AJ &  AJ & DET & N & N \\
for the POS & AV & & & & AV & & & \\
of each word & PN & & & & PP & & & \\ \hline
Phrases & what & information & is & available & about & the &
 \multicolumn{2}{l|}{{\it appears in sentences 49.}} \\
repeated & AJ & N & V & AJ & PP &
 DET &  & \\
  \cline{2-9}
within the & & & & & & the & AS/400 & system. \\
discourse & \multicolumn{5}{l}{{\it appears in sentences 6,
109, 115.}} & DET & N & N \\ \hline
POS & PN & N & V & AJ & PP & DET & N & N \\
          & AJ & & & &  & & & \\ \hline
\end{tabular}

\smallskip

N={\it noun}\ \
PN={\it pronoun}\ \
V={\it verb}\ \
AJ={\it adjective}\ \
AV={\it adverb}\ \
CJ={\it conjunction}\ \
PP={\it preposition}\ \
DET={\it determiner}
}
\end{flushleft}
\end{table*}
For example, the 39th sentence in the same chapter contains
``As you can see,'' as shown in Figure \ref{Fig-PEGInfo-assign1}.
\begin{figure*}[htbp]
\begin{quote}
\begin{quote}
{\normalsize \baselineskip=1.5\normalbaselineskip
{\sl As you can see, the help display provides additional information about
the menu options \vspace{1mm}\\
available, as well as a list of related topics.}}
{\footnotesize
\baselineskip=0.9\normalbaselineskip
\begin{verbatim}
((DECL (SUBCL  (CONJ   "as")
               (NP     (PRON*  "you" ("you" (SG PL))))
               (AUXP   (VERB*  "can" ("can" PS)))
               (VERB*  "see" ("see" PS))
               (PUNC   ","))
       (NP     (DETP   (ADJ*   "the" ("the" BS)))
               (NP     (NOUN*  "help" ("help" SG)))
               (NOUN*  "display" ("display" SG)))
       (VERB*  "provides" ("provide" PS))
        :       :       :
\end{verbatim}
\baselineskip=1.8\normalbaselineskip}
\end{quote}
\end{quote}
\caption{Thirty-ninth sentence of Chapter 6 and a part of its parse}
\label{Fig-PEGInfo-assign1}
\end{figure*}
The sentences that contain some words in common with sentence
(2.1) provide information that is very useful for deriving a correct
parse of the sentence.
Table \ref{Table-DiscInfo-assign} also shows that the parts of
speech
(POS) for most words in sentence (2.1) can be derived from
words repeated in other sentences in the same chapter.
In this table, the uppercase letters below the top sentence indicate the
parts of speech that can be assigned to the words above.
Underneath the candidate part of speech, repeated phases in other
sentences are presented along with the part of speech of each word
in those sentences;
thus, the first word of sentence (2.1), ``As,'' can be a conjunction,
an adverb, or a preposition, but complete parses of the 39th and 175th
sentences indicate that in this discourse the word is used as a
conjunction when it is used in the phrase ``As you can see.''

\begin{figure}[htbp]
 \begin{center}
 \epsfile{file=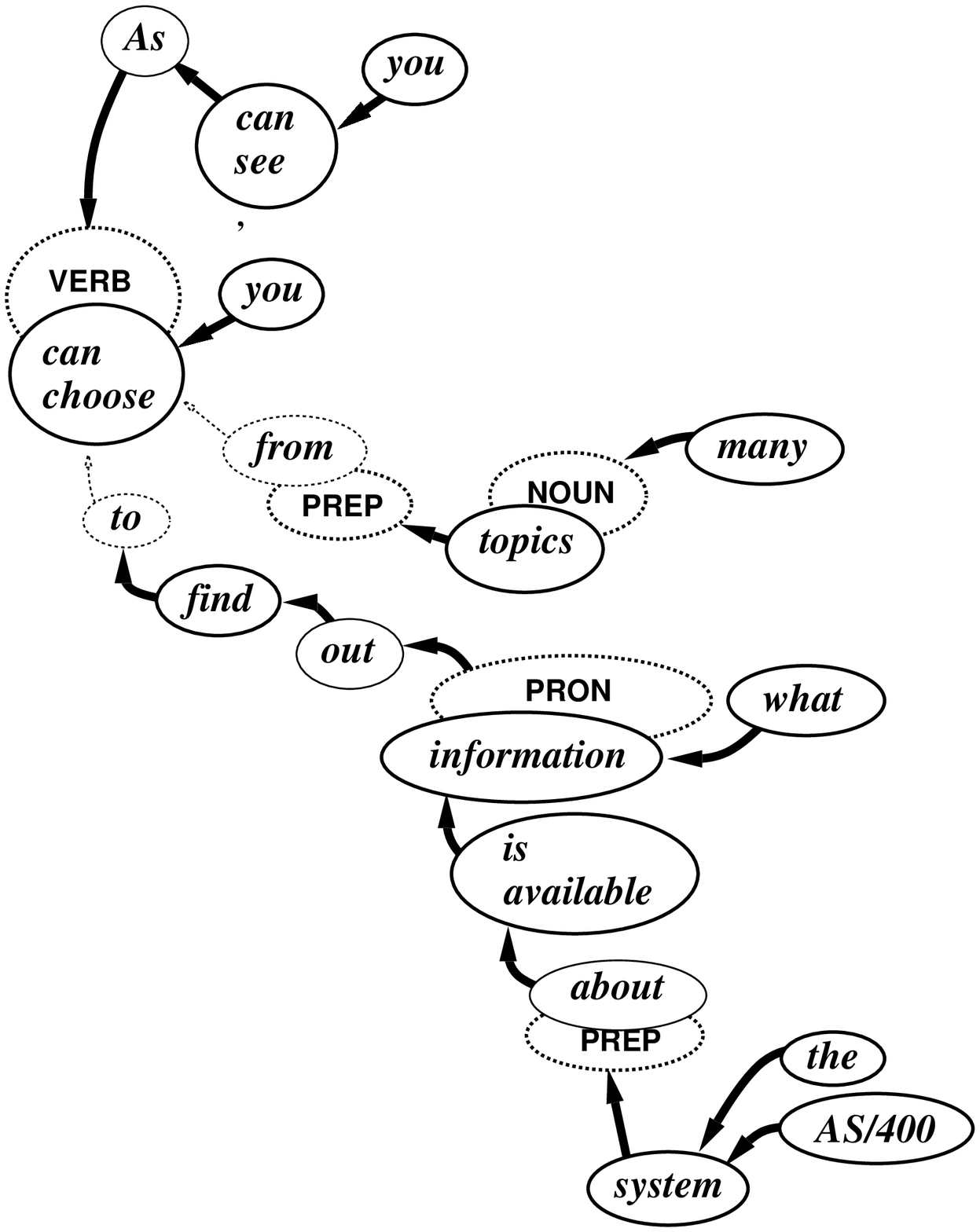,width=74mm}
 \end{center}
 \caption{Constructing a dependency structure by combining dependencies
existing within phrases that occur in other sentences of the same chapter}
 \label{PRECOMPLETE-DS}
\end{figure}

Furthermore, information on the dependencies among most words in
sentence (2.1) can be extracted from phrases
repeated in other sentences in the same chapter, as shown in Figure
\ref{PRECOMPLETE-DS}.\footnote{Thick arrows indicate dependencies
extracted from the discourse information.}

\section{Implementation}

\subsection{Algorithm}

As we showed in the previous section,
information that is very useful for obtaining correct parses of
ill-formed sentences is provided by complete parses of other sentences
in the same discourse in cases where a parser cannot construct a parse
tree by using its grammar rules.
In this section, we describe an algorithm for completing incomplete
parses by using this information.

The first step of the procedure is to extract from an input text
discourse information that the system can refer to in the next step in
order to complete incomplete parses.
The procedure for extracting discourse information is as follows:

\begin{enumerate}
\item Each sentence in the whole text given as a discourse is processed
by a syntactic parser.  Then, except for sentences with incomplete
parses and multiple parses, the results of each parse are stored as
discourse information.
To be precise, the position and the part of speech of each instance of
every lemma are stored along with the lemma's modifiee-modifier
relationships with other content words extracted from the parse data.
Table \ref{Table-Dframe} shows an example of such information.
\begin{table*}[ht]
\caption{Discourse information on modifiees and modifiers of a noun ``cursor''}
\label{Table-Dframe}
\begin{center}
{\small
\begin{tabular}{|c|c|l|} \hline
 \multicolumn{3}{|c|}{Modifiers} \\ \hline
 POS & Relation & Word ({\tt CFRAME}s \ \  preference value)
\\ \hline
 Noun & of & display ({\tt CFRAME106873} 0.1)  \\ \cline{2-3}
      & in & protected area ({\tt CFRAME106872} 1)  \\ \cline{2-3}
      & to & left ({\tt CFRAME106407} 0.1) \ \
right({\tt CFRAME106338} 0.1) \\ \cline{2-3}
      & {\tt DIRECT} & position ({\tt CFRAME106405} 1) \\
 \hline
 Adjective & up & line ({\tt CFRAME106295} 0.1)  \\ \cline{2-3}
	   & {\tt DIRECT} &
 your ({\tt CFRAME106690 CFRAME106550} 2) \\ \hline  \hline
 \multicolumn{3}{|c|}{Modifiees} \\ \hline
 POS & Relation & Word ({\tt CFRAME}s \ \  preference value) \\ \hline
 Verb & with & play ({\tt CFRAME106928} 0.1) \ \
 be ({\tt CFRAME106927} 0.1) \\ \cline{2-3}
      & up   & move ({\tt CFRAME106688} 1) \\ \cline{2-3}
      & {\tt SUBJ} & stop ({\tt CFRAME106572} 1) \ \
 reach ({\tt CFRAME106346} 1) \ \ move ({\tt CFRAME106248} 1) \\
\cline{2-3}
      & {\tt OBJ} & move ({\tt CFRAME106402 CFRAME106335 CFRAME106292} 3)
confuse ({\tt CFRAME106548} 1) \\ \cline{2-3}
      & {\tt RECIPIENT} & move ({\tt CFRAME106304} 1) \\ \hline
\end {tabular}
}
\end{center}
\end{table*}
In this table, {\tt CFRAME\verb*+      +} indicates an instance of {\it
cursor} in the discourse; information on the position
and on the whole sentence can be extracted from each occurrence of {\tt
CFRAME}.
In accumulating discourse information, a score of 1.0 is awarded for
each definite modifiee-modifier relationship.  A lower score, 0.1, is
awarded for each ambiguous modifiee-modifier relationship, since such
relationships are less reliable.

\item When all the sentences have been parsed, the discourse information
is used to select the most preferable candidate for sentences
with multiple possible parses, and the data of the selected
parse are added to the discourse information.
\end{enumerate}

After all the sentences
except the ill-formed sentences that caused incomplete parses have
provided data for use as discourse
information, the parse completion procedure begins.

The initial data used in the completion procedure are a set of partial
parses generated by a bottom-up parser as an incomplete parse tree.
For example, the PEG parser generated three partial parses for
sentence (2.1), consisting of ``As you can see,'' ``you can choose from
many topics,'' and ``to find out what information is available about the
AS/400 system,'' as shown in Figure \ref{Fig-PEG-output}.
Since partial parses are generated by means of grammar rules in a
parser, we decided to restructure each partial parse and unify them
according to the discourse information, rather than construct the
whole parse tree from discourse information.

The completion procedure consists of two steps:

\subsubsection*{Step 1: Inspecting each partial parse and restructuring
it on the basis of the discourse information}
For each word in a partial parse, the part of speech and the
modifiee-modifier relationships with other words are inspected.
If they are different from those in the discourse information, the partial
parse is restructured according to the discourse information.

\begin{figure}[t]
 \begin{center}
 \psbox[width=74mm]{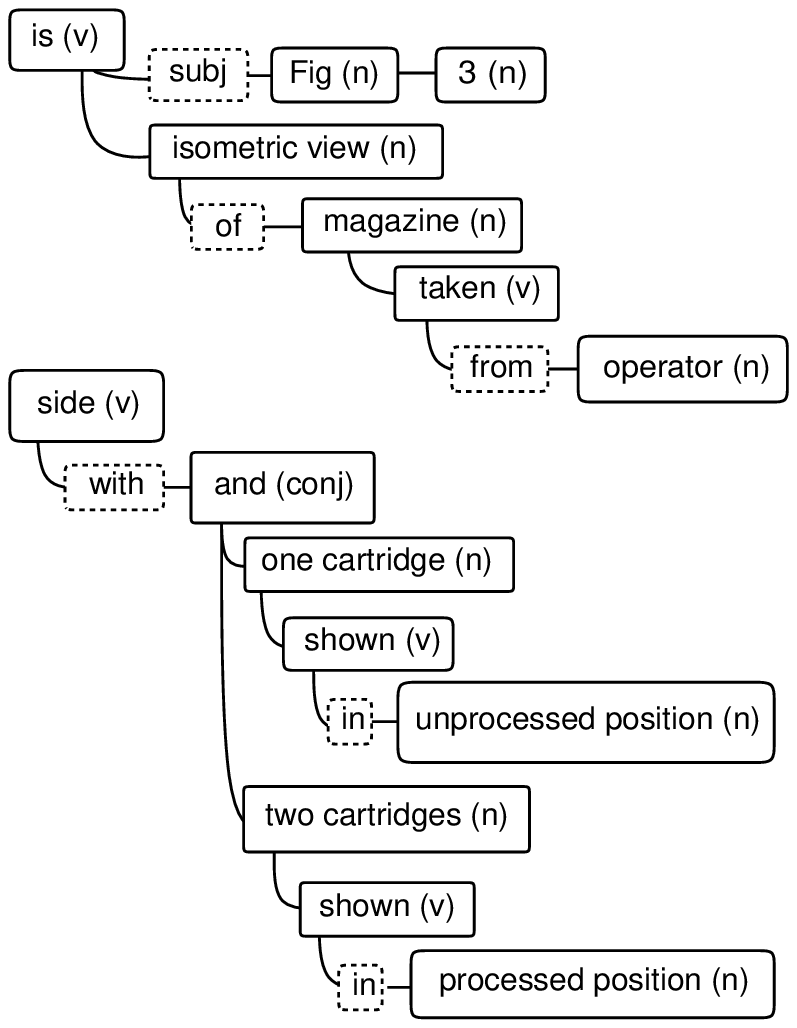}
 \end{center}
\caption{Example of an incomplete parse by the ESG parser}
\label{Fig-ESG-output}
\end{figure}
\begin{figure}[t]
 \begin{center}
 \psbox[width=84mm]{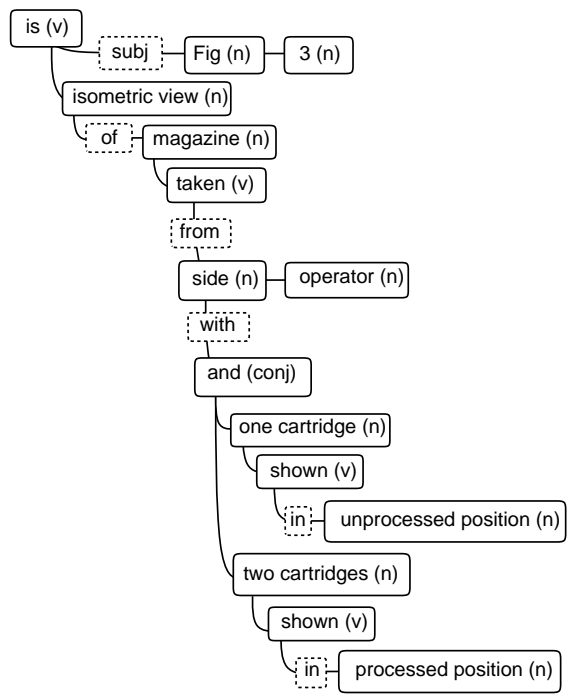}
 \end{center}
\caption{Example of a completed parse}
\label{Fig-recovered-output}
\end{figure}

For example, Figure \ref{Fig-ESG-output} shows an incomplete parse of
the following sentence, which is the 43rd sentence
in a technical text that consists of 175 sentences.\footnote{This
structure resulting from an incomplete parse does not indicate that the
grammar of the parser lacks a rule for handling a possessive case
indicated by an apostrophe and an s.  When the parser fails to generate
a unified parse, it outputs partial parses in such a manner that fewer
partial parses cover every word in the input sentence.}
\begin{description}
\item[(3.1)]{\sl Fig. 3 is an isometric view of the magazine taken from
the operator's  side with one cartridge shown in an unprocessed position
and two  cartridges shown in a processed position.}
\end{description}
In the second partial
parse, the word ``side'' is analyzed as a verb.
The same word appears fifteen times in the discourse information
extracted from well-formed sentences, and is analyzed as a noun
every time it appears in complete parses;
furthermore, there are no data on the
noun ``operator'' modifying the verb ``take'' through the preposition
``from,'' while there is information on the noun ``operator's'' modifying
the noun ``side,'' as in sentence (3.2), and on the noun ``side''
modifying the verb ``take,'' as in sentence (3.3).
\begin{description}
\item[(3.2)]{\sl In the operation of the invention, an operator loads
cartridges into the magazine from \underline{the operator's side} as
seen in Figs. 3 and 12.} (151st sentence)
\item[(3.3)]{\sl Fig. 4 is an isometric view of the magazine \underline{taken
from the machine side} with one cartridge shown in the unprocessed position and
two cartridges  shown in the processed position.} (44th sentence)
\end{description}
Therefore, these two partial parses are restructured by changing
the part of speech of the word ``side'' to noun, and the
modifiee of the noun ``operator'' to the noun ``side,'' while at the same
time changing the modifiee of the noun ``side'' to the verb
``take.''
As a result,
a unified parse is obtained, as shown in Figure \ref{Fig-recovered-output}.

\subsubsection*{Step 2: Joining partial parses on the basis of the
discourse information}
If the partial parses are not unified into a single structure in the
previous step,
they are joined together on the basis of the discourse information
until a unified parse is obtained.

Partial parses are joined as follows:

First, the possibility of joining the first two partial parses
is examined, then, either the unification of the first two parses or
the second parse is examined to determine whether it can be joined to
the third parse,
then the examination moves to the next parse, and so on.

Two partial parses are joined if the root (head node) of either parse
tree can modify a node in the other parse without crossing the
modification of other nodes.

To examine the possibility of modification, discourse
information is applied at three different levels.
First, for a candidate modifier and modifiee,
an identical pattern containing the modifier word and the modifiee word
in the same part of speech and in the same relationship is searched for
in the discourse information.
Next, if there is no identical pattern, a modification pattern with a
synonym \cite{COLLINS:1984} of the node on one side is searched for in
the discourse information.
Then,
if this also fails, a modification pattern containing a word that has
the same part of speech as the word on one side of the node is
searched for.

Since the discourse information consists of modification patterns
extracted from complete parses, it reflects the grammar rules of the
parser, and a matching pattern with a part of speech rather than an
actual word on one side can be regarded as a relaxation rule, in the
sense that syntactic and semantic constraints are less restrictive than
the corresponding grammar rule in the parser.

These matching conditions at different levels are applied in such a
manner that partial parses are joined through the most preferable nodes.

\subsection{Results}

We have implemented this method on an English-to-Japanese machine
translation system called Shalt2 \cite{Takeda:1992}, and conducted
experiments to evaluate the effectiveness of this method.
Table \ref{Table-result} gives the result of our experiments on
two technical documents of different kinds,
one a patent document (text 1), and the other a computer manual
(text 2).
Since text 1 contained longer and more complex sentences than text 2,
our ESG parser failed to generate unified parses more often in
text 1; on the other hand, the frequency of morphologically identical
words and collocation patterns was higher in text 1, and our method was
more effective in text 1.
In both texts, the discourse information provided enough information to
unify partial parses of an incomplete parse in more than half of the
cases.
However, the resulting unified parses were not always correct.
Since sentences with incomplete parses are usually quite long and
contain complicated structures, it is hard to obtain a perfect analysis
for those sentences.
Thus, in order to evaluate the improvement in the output translation
rather than the improvement in the rate of success in syntactic
analysis, in which only perfect analyses are counted, we compared output
translations generated with and without the application of our method.
When our method was not applied,
partial parses of an incomplete parse were joined by means of some
heuristic rules such as the one that joins a partial parse with ``NP''
in its root node to a partial parse with ``VP'' in its root node, and
the root node of the second partial parse was joined to the last node of
the first partial parse by default.
When the discourse information did not provide enough information to
unify partial parses with the application of our method,
the heuristic rules were applied.
In such cases the default rule of joining the root node of the second
partial parse to the last node of the first partial parse was mostly
applied, since the least restrictive matching patterns in our method
were similar to the heuristic rules.
Thus, the system generated a unified parse for each sentence regardless of
the discourse information, and we compared the output translations
generated with and without the application of our method.
The results are shown in Table \ref{Table-result}.
The translations were compared by checking how well the output Japanese
sentence conveyed the meaning of the input English sentence.
Since most unified parses contained various errors, such as incorrect
modification patterns and incorrect parts of speech assigned to some
words, fewer errors generally resulted in better translations, but
incorrect parts of speech resulted in worse translations.

\begin{table*}
\caption{Results of completing incomplete parses on the basis of
discourse information}
\label{Table-result}
\begin{center}
{\small
\begin{tabular}{|c|c|c|c|} \hline
\multicolumn{2}{|c|}{} & Text1 & Text2 \\ \hline
\multicolumn{2}{|c|}{Number of sentences in discourse} & 175 & 354 \\ \hline
\multicolumn{2}{|c|}{Incomplete parses}	& 32 & 31\\ \hline \hline
\multicolumn{2}{|c|}{Unified into a single parse} & 18 (56.3\%) & 17
(54.8\%) \\ \hline
Improvement & Better & 7 & 7 \\ \cline{2-4}
in &	Even & 10 & 7\\ \cline{2-4}
translation &	Worse & 1 & 3 \\ \hline
\multicolumn{2}{|c|}{Partially joined or restructured} & 12 (37.5\%) &
8 (25.8\%) \\ \hline
Improvement & Better & 4 & 2 \\ \cline{2-4}
in &	Even & 7 & 3 \\ \cline{2-4}
translation & Worse & 1 & 3 \\ \hline
\multicolumn{2}{|c|}{Not changed} & 2 (6.3\%) & 6 (19.4\%) \\ \hline
\end {tabular}
}
\end{center}
\end{table*}

\section{Conclusion}

We have proposed a method for completing partial parses of
ill-formed sentences on the basis of information extracted from complete
parses of well-formed sentences in the discourse.
Our approach to handling ill-formed sentences is fundamentally different
from previous ones in that it reanalyzes the part of speech and
modifiee-modifier relationships of each word in an ill-formed sentence
by using information extracted from analyses of other sentences in the
same text, thus, attempting to generate the analysis most appropriate
to the discourse.
The results of our experiments show the effectiveness of this method;
moreover, implementation of this method on a machine translation system
improved the accuracy of its translation.
Since this method has a simple framework that does not require any
extra knowledge resources or inference mechanisms, it is robust and
suitable for a practical natural language processing system.
Furthermore, in terms of the turn-around time (TAT) of the whole
translation procedure,
the improvement in the parses achieved by using this method along with
other disambiguation methods involving discourse information, as shown
in another paper \cite{Nasukawa:1995}, shortened the TAT in the late
stages of the translation procedure, and compensated for the extra TAT
required as a result of using the discourse information, provided the
size of the discourse was kept to between 100 and 300 sentences.

In this paper, the term ``discourse'' is used as a set of words in a
text together with the usage of each of those words in that text --
namely, a part of speech and modifiee-modifier relationships with other
words.
The basic idea of our method is to improve the accuracy of sentence
analysis simply by maintaining consistency in the usage of
morphologically identical words within the same text.
Thus, the effectiveness of this method is highly dependent on the source
text, since it presupposes that morphologically identical words are
likely to be repeated in the same text.
However, the results have been encouraging at least with technical
documents such as computer manuals, where words with the same lemma are
frequently repeated in a small area of text.
Moreover, our method improves the translation accuracy, especially for
frequently repeated phrases, which are usually considered to be
important, and leads to an improvement in the overall accuracy of the
natural language processing system.

\section*{Acknowledgements}

I would like to thank Michael McDonald for invaluable help in
proofreading this paper.
I would also like to thank
Taijiro Tsutsumi, Masayuki Morohashi, Koichi Takeda, Hiroshi Maruyama,
Hiroshi Nomiyama, Hideo Watanabe, Shiho Ogino, Naohiko Uramoto,
and the anonymous reviewers for their comments and suggestions.


\begin{thebibliography}{999}

\bibitem[\protect\citename{Douglas and Dale}1992]{Douglas:1992}
Douglas, S. and Dale, R.
\newblock 1992.
\newblock Towards Robust PATR.
\newblock In {\em Proceedings of COLING-92}.

\bibitem[\protect\citename{Gale \bgroup et al.\egroup }1992]{Gale:1992b}
Gale, W.A., Church, K.W., and Yarowsky, D.
\newblock 1992.
\newblock One Sense per Discourse.
\newblock In {\em Proceedings of the 4th DARPA Speech and Natural
Language Workshop}.

\bibitem[\protect\citename{Jensen \bgroup et al.\egroup }1992]{Jensen:1983}
Jensen, K., Heidorn, G.E., Miller, L.A. and Ravin, Y.
\newblock 1983.
\newblock Parse Fitting and Prose Fixing: Getting a Hold on
Ill-Formedness.
\newblock {\em Computational Linguistics}, Vol. 9, Nos. 3-4.

\bibitem[\protect\citename{Jensen}1992]{Jensen:1992}
Jensen, K.
\newblock 1992.
\newblock PEG: The PLNLP English Grammar.
\newblock {\em Natural Language Processing: The PLNLP Approach},
 K. Jensen, G. Heidorn, and S. Richardson, eds., Boston, Mass.: Kluwer
Academic Publishers.

\bibitem[\protect\citename{McCord}1991]{McCord:1991}
McCord, M.
\newblock 1991.
\newblock The Slot Grammar System.
\newblock {\em IBM Research Report}, RC17313.

\bibitem[\protect\citename{Nasukawa}1993]{Nasukawa:1993}
Nasukawa, T.
\newblock 1993.
\newblock Discourse Constraint in Computer Manuals.
\newblock In {\em Proceedings of TMI-93}.

\bibitem[\protect\citename{Nasukawa}1995]{Nasukawa:1995}
Nasukawa, T.
\newblock 1995.
\newblock Shallow and Robust Context Processing for a Practical MT
System.
\newblock To appear in {\em Proceedings of IJCAI-95 Workshop on
``Context in Natural Language Processing.''}

\bibitem[\protect\citename{Richardson and Braden-Harder}1988]{Richardson:1988}
Richardson, S.D. and Braden-Harder, L.C.
\newblock 1988.
\newblock The Experience of Developing a
Large-Scale Natural Language Text Processing System: CRITIQUE.
\newblock In {\em Proceedings of ANLP-88}.

\bibitem[\protect\citename{Takeda \bgroup et al.\egroup }1992]{Takeda:1992}
Takeda, K., Uramoto, N., Nasukawa, T., and Tsutsumi, T.
\newblock 1992.
\newblock Shalt2 -- A Symmetric Machine Translation System with Conceptual
Transfer.
\newblock In {\em Proceedings of COLING-92}.

\bibitem[\protect\citename{IBM}1992]{IBM:1992}
IBM
\newblock 1992.
\newblock {\em IBM Application System/400 New User's Guide Version 2}.
\newblock IBM Corp.

\bibitem[\protect\citename{Collins}1984]{COLLINS:1984}
COLLINS
\newblock 1984.
\newblock {\em The New Collins Thesaurus}.
\newblock Collins Publishers, Glasgow.

\end{thebibliography}
\end{document}